# A Biomimetic Model of the Outer Plexiform Layer by Incorporating Memristive Devices

A. Gelencsér[*+], T. Prodromakis[+], C. Toumazou[+] and T. Roska[*]

[*] *Interdisciplinary Technical Sciences Doctoral School, Pázmány Péter Catholic University, Budapest, Hungary*

[+] *Centre for Bio-inspired Technology, Department of Electrical and Electronic Engineering, Imperial College London, London, UK*

**In this paper we present a biorealistic model for the first part of the early vision processing by incorporating memristive nanodevices. The architecture of the proposed network is based on the organisation and functioning of the outer plexiform layer (OPL) in the vertebrate retina. We demonstrate that memristive devices are indeed a valuable building block for neuromorphic architectures, as their highly non-linear and adaptive response could be exploited for establishing ultra-dense networks with similar dynamics to their biological counterparts. We particularly show that hexagonal memristive grids can be employed for faithfully emulating the smoothing-effect occurring at the OPL for enhancing the dynamic range of the system. In addition, we employ a memristor-based thresholding scheme for detecting the edges of grayscale images, while the proposed system is also evaluated for its adaptation and fault tolerance capacity against different light or noise conditions as well as distinct device yields.**

Over the past years, the performance and efficiency of biological systems has inspired many researchers and engineers, giving birth to the emerging fields of biomimetics [1] and bio-inspiration [2]. The human retina is anything but a simple passive relay station, as it pre-processes and compresses all sensed information through an immensely complex neuronal network that on average contains 4.6 million cones, 92 million rods [3] and 1 million ganglion [4]. If we take into account the remaining bipolar, horizontal and amacrine cells and the fact that they are highly interconnected we get a parallel network of grand complexity. This complexity is further elucidated in recent studies where evidence is provided that at least ten parallel signals arise from a single visual point [5] and we get a much more accurate view from the functioning of the vertebrate retina. In fact it is nowadays believed that although biological systems are based on relatively primitive elements, it is this naturally occurring interconnection complexity that facilitates higher order functioning. In this paper we specifically focus on the connection of the sensory and consecutive system of the retina, which is the first and common step in the visual information flow. These connections take place on the outer plexiform layer (OPL) in the vertebrate retina; form a highly dynamic system that enables the smoothing of optical inputs and thus catalyzes the enhancement of the retina's dynamic range, while the different parallel channels emerge after this point.

Our approach is distinct from the already existing models, described in [6]-[22]. The use of time variant resistors to model biological network raises principal and practical concerns. Our approach alleviates these issues by employing the latest biological knowledge [5], and an emerging nanoscale device that is used as a more adequate synapse emulator [23], the memristor. This device exhibits a highly non-linear dynamic behaviour, which along its infinitesimal dimensions serves as an excellent building block for facilitating practical realisations of the highly complex synaptic networks constituting the OPL. We further expand this approach by utilising a memristive-based thresholding scheme for performing edge-detection. Finally, we demonstrate that this platform exhibits similar attributes to naturally occurring systems such as noise resilience, self-adaptation and fault tolerance.

**Biological Network.** Neurobiologists have identified five major classes of neurons, which in the mammalian retina are divided into about 60 anatomically different types of cells [24], distributed across 7 distinct layers. We particularly focus on the outer plexiform layer and eminently to the synaptic connections and corresponding signal propagation through these.

**The Mammalian Retina.** Fig. 1 shows a schematic cross-section of the mammalian retina. Five major neuron types are denoted: photoreceptors, bipolar, ganglion, horizontal and amacrine cells. Manifold approaches exist for studying the retina, nevertheless in this paper we tend to follow the information flow. In the vertical direction, photoreceptors detect the incidental light through the night vision rods and the cones that are responsible for the daylight and colour vision. These are in direct contact with bipolar cells, which relay the visual information in the form of membrane depolarisation or hyperpolarisation to the ganglion cells. The axons of the latter cells propagate the pre-processed visual information towards the brain, essentially forming the optic nerve, with ON cells depolarizing and OFF cells hyperpolarizing in accordance to the corresponding impulses.

On the other hand, the information flow in the lateral direction is mainly mediated by the horizontal and amacrine cells, as shown in Fig. 1. The horizontal cells build synaptic connections with the photoreceptors and the input of bipolar cells in the outer plexiform layer (OPL). At the same time, the interconnections between the output of bipolar cells, amacrine cells and retinal ganglion cells take place in the inner plexiform layer (IPL).

The cells in the retina use two modes to transmit their signals. The photoreceptors, bipolar and horizontal cells as well as some of the amacrine cells when exposed to a stimulus produce graded changes in their membrane potential [25]-[26], permitting a fast and continuous signal flow. On the other hand, all of the ganglion cells and some amacrine cells produce action potentials [27]. And although the attainable propagation velocity is much slower, it is rather robust, allowing the information to travel over long-distances.

**Outer Plexiform Layer.** The horizontal cells play an important role in the synaptic interactions within the OPL. These cells are electrically interconnected through gap junctions, as shown in Fig. 2(a) forming a complex lateral network. These cells are not just receiving information from the photoreceptors; a feedback mechanism [28] exists that pools the information from the horizontal cell network over a wide spatial area of the OPL that adjusts the photoreceptors gain. In the case where all the photoreceptor cells are evenly illuminated, then the horizontal cells will be accordingly stimulated evenly and all nodes in the OPL will essentially be equipotential. Therefore there is no current flowing between them. If there is however a small spot light stimuli, then the membrane potential of the not affected horizontal cells will differ, allowing the conduction of electrical currents between these cells; essentially shunting the responses of the illuminated ones. This has two main ramifications: firstly the photoreceptor's response is proportional to the ratio between its photo input and the local average of the surround region of the retina and secondly this local circuit provides the bipolar cells with a center-surround organisation.

The current flow between the horizontal cells diminishes the gradient between the adjacent photo inputs, and essentially facilitates a local Gaussian filtering for smoothing the input image. The smoothed visual signal then splits into two separate channels of information [29], detecting the lighter and the darker objects of the background via ON and OFF pathways. The combination of these pathways creates simultaneous contrast (light and dark boundaries) of visual objects (center-surround receptive field) in deeper layers [30]. The amacrine cells in the IPL play a critical role in detecting the speed and orientation of motion [31]-[32] as well as determining spatial colour vision or nocturnal vision [33].

We scrutinise the architecture of the synaptic interconnections in the OPL to create our model. These interconnects occur in the lower part of the cone (the cone pedicle), often called triads. The central element of a triad is an invaginating bipolar dendrite, originating from an ON bipolar cell, while the lateral elements are invaginating horizontal cell dendrites or axons. The triad also comprises of bipolar dendrites, originating from OFF bipolar cells that establish basal junctions at the bottom of the cone pedicle [34], as depicted in Fig. 2(b).

**Retinomorphic Memristive Grids.** The photoreceptor cells of more developed species are typically tessellated in an hexagonal manner, as can be seen in Fig. 2 of [35], with the layers underneath following this same hexagonal arrangement for ensuring an effective spatial usage [36]. Modelling the triads occurring in the OPL, necessitates the employment of elements with infinitesimal dimensions that are capable of dynamically-adapting their weight in accordance to the current flowing through these. Here we describe how the attributes of the recently discovered memristor can be utilised in this particular bio-inspired system.

**Memristive Components.** The recent nanoscale implementation of the memristor by Hewlett-Packard (HP) [37] and particularly its application as a synapse emulating device (demonstrated in [23] and also [38]-[39]), has created a lot of excitement within the neuromorphic community. The device was theoretically predicted in 1971 by Leon Chua in his seminal paper [40] and was postulated as a variable resistor with a state-dependent dynamic response:

$$v(t) = M(q)i(t) \tag{1}$$

where $M(q)$ is the charge-dependant resistance, or more appropriately the memristance. At the lowest approximation where a linear ionic dopand drift is considered, the memristance will take any value between two limiting resistive states: high-resistive state (HRS) $R_{off}$ and a low-resistive state (LRS) $R_{on}$ described as:

$$M(q) = R_{on}\frac{w(t)}{D} + R_{off}\left(1 - \frac{w(t)}{D}\right) \tag{2}$$

where $w(t)$ is the width of the doped region and $D$ is the total thickness of the bi-layer or active region of the device. $R_{on}$ is essentially the net resistance of the device when the active region is completely doped $(w = D)$ and $R_{off}$ when the opposite occurs $(w\rightarrow0)$. More information on the intrinsic properties of the models we employed can be found in [41] and [42] (also see methodology section).

The strength (conductance) modulation of a single memristive element does not only depend on the charge that has flown through the device but also on the direction this has flown. This polarity dependence is however an undesirable feature for the proposed OPL model and a typical way around this is the utilisation of two devices in series with opposing polarities, as shown in Fig. 3, also denoted as a memristive fuse [43]. The beauty of this combination is that it overall maintains the non-linear relationship between the time integrals of current and voltage, without demonstrating any polarity dependence. This is a very useful characteristic for preserving the edges on the input image, with more details given in section 4.1.

**Biomimetic OPL Triads.** Any visual input is captured via the photoreceptor cells of the retina. As this is out of the scope of this paper, we opted not emulating the exact functioning of these receptors, rather the effect in translating any light stimuli into an appropriate current bias. We thus represent the cells' signalling with equivalent voltage sources. Clearly, this approach is only applicable for colourless image inputs.

As illustrated in Fig. 4, every voltage source connects via a resistor to a discrete node of an hexagonal memristive network. The value of this resistance is comparable to the initial memristive states ($1k\Omega$), since the smaller this resistance is the larger the current in the underlying node will be. It therefore serves as a control parameter of the time-evolution of our system. These nodes are essentially representations of the triads comprising the OPL [44], with the memristive fuses employed for emulating the synaptic interconnects of the horizontal cell network. In a similar manner to Fig. 2(b), single memristors were used to emulate the bipolar cells' dendrites, while the biasing provided by the voltage sources and their series resistors resemble the activity occurring at the cone pedicles.

Here we have ascertained an hexagonal topology for complying with the biological counterpart system. Every node within the grid is linked with six neighbouring nodes through memristive fuses. This approach is essentially similar to the one employed by Carver Mead in [6] with our system being distinct in that we employed non-linear memristive devices instead of linear resistors. This option facilitates the proposed model with an inherent local Gaussian filtering functioning to any input image.

**Localised Smoothing.** In order to demonstrate the network's adaptation capacity we first examine the response of a relatively small network while subjected to an optical stimuli comprising of a large contrast. This stimuli is illustrated in the inset of Fig. 5(a) with the centre node being biased with a DC voltage of 30 mV (black pixel), while the neighbouring pixels were all grounded. This essentially establishes a large potential difference across the devices that are affiliated to this particular node (denoted as green lines in the inset A of Fig. 5(b)). Clearly all devices that link the nodes adjacent to the centre node are equipotential and thus no memristance modulation is expected to occur at any time. Nevertheless, small static potentials are established on the remaining nodes due to the time-evolution of the system. All memristive fuses that are subjected to similar biasing conditions are labelled with the same colour, as illustrated in Fig. 5(b). The larger the current flow through the adjacent nodes is the greater the state modulation rate.

This simple network essentially adapts to the applied bias through the establishment of distinct time dependent currents that modulate the corresponding states of the devices, as illustrated in Fig. 5(b). In turn, the corresponding static potentials of the nodes evolve after t=30sec in the distribution shown in Fig. 5(a), with the central, neighbouring and peripheral nodes' potentials being 21.9mV, 0.6mV and 0.4mV respectively. The time-evolution of the nodes' potential is further illustrated in S.1.

This response can be analysed by observing the current flows through the memristive fuses adjacent to the centre node. Node N is surrounded by 6 adjacent nodes and links to these via memristive fuses $M_1$, $M_2$, $M_3$, $M_4$, $M_5$ and $M_6$. The input voltage bias $V_{p(i,j)}$ is conveyed to a stimulating current through a series resistor $R_{i,j}$. Let us consider the voltage at node N to be $V_{out}$, while the corresponding voltages at the six neighbouring nodes being $V_1$, $V_2$, $V_3$, $V_4$, $V_5$ and $V_6$ respectively. Kirchhoff's Current Law reads:

$$I_{in} = \sum_{i=1}^{6} I_i \tag{3}$$

and according to Ohm's Law :

$$\frac{V_{p(i,j)} - V_{out}}{R_{i,j}} = \sum_{i=1}^{6} \frac{V_{out} - V_i}{M_i} \tag{4}$$

Rearranging equation (4) reads:

$$V_{out} = \frac{V_{p(i,j)} + R_{i,j} \sum_{i=1}^{6} \frac{V_i}{M_i}}{1 + R_{i,j} \sum_{i=1}^{6} \frac{1}{M_i}} \tag{5}$$

Therefore, the corresponding output voltage $V_{out}$ at node N at any time instance is related to the weighted average of the potentials established in the surrounding nodes.

**Biomimetic Outer Plexiform Layer.** Typical resting potential of neuron cells is -65mV, with the photoreceptors resting membrane potential being about -40mV. When some light stimulus is present, these cells use the absorbed photons energy to hyperpolarise their inherent membrane potential, i.e. light intensity of an optical stimulus will cause photoreceptor cells to hyperpolarise in relation to the light intensity. On the other hand a dark intensity will have an opposing effect, causing the cells to depolarise, as illustrated in Fig. 6.

In this approach the stimulating potential resulting from photoreceptor cells being either hyperpolarised or depolarised was modulated arbitrarily through the employed voltage sources $V_{p(i,j)}$. Every pixel of an input image was represented with a corresponding voltage level, as depicted in the inset of Fig. 6. The pixel intensity of all stimuli was set on a 16-level grayscale, where value 0 represent the highest possible light intensity and 15 corresponds to a dark current. Thus, all optical inputs were transcribed into a 2D vector of biasing voltages that through a serial resistor establish the OPL's stimulating currents. In a similar manner, the recorded potentials of the memristive OPL nodes are monitored as the network is adjusting the corresponding memristive weights and is transcribed back to matching pixel intensities, for obtaining some meaningful data. We should note here that as the employed depolarisation potentials are in the mV range, these will provoke a relatively slow state-modulation of the memristive network. In consequence, the system evolvement is better observed in the time period of a few seconds rather in milliseconds, which however can be adjusted by modifying the stimuli amplitudes. Finally, throughout all evaluations we utilised a cartoon that comprises of clear edges along with areas of finer details and a simplified picture of a Rubik's Cube that has clear patterns.

**Smoothing and local Gaussian filtering.** Most of the existing edge-detection schemes demonstrate various issues when the optical input is distorted [45]. To address these, Gaussian filtering was proposed for suppressing noise through the smoothing (blurring) of the image. A one dimensional Gaussian function can be described as:

$$g(x) = \frac{1}{\sqrt{2\pi}\sigma} e^{-\frac{x^2}{2\sigma^2}} \tag{6}$$

where $x$ and $y$ are the distance from the origo and $\sigma$ is the standard deviation of the Gaussian distribution, which could also be expanded into two-dimensions through:

$$g(x,y) = \frac{1}{2\pi\sigma^2} e^{-\frac{x^2+y^2}{2\sigma^2}} \tag{7}$$

Nevertheless, a uniform Gaussian blur across the whole image can cause the displacement of edges, the vanishing of less intense edges along with the creation of edge artefacts [46]-[47]. The occurrence of this effect however is diminishable, in the case of local Gaussian filtering [48]. In our approach, memristive dynamics are employed for achieving this performance intrinsically. The filtering variance is dynamically adapting to the local variance of the image and the smoothing alleviates any non-uniformities.

The amplitude of the current flowing through any memristive fuse in the OPL depends on the weighted sum of the current flow through the neighbouring cells. If the potential difference between adjacent nodes is high, then the current flow gets higher; if the difference is however low, then the current flow is significantly less. In other words, a clear edge on the input image causes a big intensity gradient and consequently a faster memristance modulation. Such memristive fuses drift towards a HRS in a faster manner than the neighbouring cells and the lateral current flow from the high pixel gradient essentially diminishes. In consequence, the edge is preserved while the smoothing effect is vigorously decreased. On the other hand, if the contrast between adjacent pixels is low, the corresponding memristance change will be significantly slower. In this case, this particular memristive fuse will allow a larger lateral current flow and thus the resulting image will be more homogeneous over this area. Hence, small intensity variations on an image tend to smooth out, meaning that added noise is also annihilated, proving the feasibility of the proposed memristive-based local Gaussian filtering.

This is particularly demonstrated through the examples depicted in Fig. 7(a) and (7c), with Fig. 7(a) being the original input image and Fig. 7(c) a distorted version of the original image with additive white noise with a Gaussian distribution ($\mu$=0 and $\sigma = 0.3$). In both cases the smoothing caused by this network after $t$=30 seconds is illustrated in Fig. 7(b) and Fig. 7(d) respectively. These figures depict the static voltages measured at the OPL's nodes after being transcribed back to the corresponding grayscale intensities in accordance with the scale shown in the inset of Fig. 6. By observation, the two figures do not show any considerable difference and this was quantified in terms of grade intensity mismatch to be approximately 3%, as shown in Fig. 7(e). In both cases, the smoothed versions preserved the main edges, while wherever there was an insignificant intensity gradient or small intensity variation caused by the addition of noise, the system smoothed this out. The smoothing transient of both cases is also illustrated dynamically in S.2a and S.2b.

**Edge Detection.** The intensity contrast between adjacent pixels imposes the biasing of the underlying memristive fuses with corresponding potential differences and as such an edge can easily be detected by monitoring the outgoing current flow at the OPL nodes. In the counterpart biological system, this information is conveyed through the dendrites of ON/OFF bipolar cells to the IPL. It is important to mention that this edge detection phenomenon is a part of the early vision processing, while the main output of the edge information occurs later in the retina. Here, we utilize two approaches: (1) we employ single memristors in the output stage of the OPL to facilitate a resistive thresholding scheme and (2) we monitor the state variance of the OPL memristive fuses.

At any given time, the relative memristance change, both in the memristive fuses and the single devices, is a measure of the current flowing through these devices. As the system evolves, the devices associated with nodes that are exposed to large potentials, i.e. in neighbouring pixels at an edge, will drift towards lower conductive states at a rate set by the overlying intensity contrast. By monitoring the transient memristance change of the devices in the OPL output nodes, we associate appropriate thresholding values for defining clear edges. All devices falling between these thresholds will thus indicate the existence of an edge. In the case where Fig. 8(a) is the source image and the threshold is bounded within $600\Omega \le M_T \le 2k\Omega$, the detected edges will correspond to what is shown in Fig. 8(b). Clearly, these thresholds could be manually adjusted for attaining more or less edge details.

In the second approach, a more elaborate thresholding scheme involves the monitoring of the states of all memristive fuses associated with a node. If at least three devices are exceeding a preset threshold, this particular node is then denoted as an edge pixel. Fig. 8(c) illustrates the edges as detected through this approach for a $M_T = 1.6k\Omega$. Supplementary videos S.3a and S.3b demonstrate the transient response of our system, Fig. 7(a) and Fig. 7(c) are respectively used to bias the memristive network. Additionally, as a figure of merit, Figs. 8(d), 8(e) and 8(f) illustrate the edges detected by employing conventional algorithms, specifically: Prewitt, Sobel and Canny.

Similarly, this method was also exploited with a Rubik's cube image, as shown in Fig. 9(a), with the memristive threshold being set at $M_T = 3k\Omega$. In this example, the smoothing process has caused some distortion on the inhibited pattern of the front sides of the cube, as this is erroneously considered as noise due to the small contrast difference existing between these single-pixel lines and their background. As a result this pattern tends to be smoothed out, as illustrated in Fig. 9(b) as well as in supplementary video S.4. Nevertheless, our model manages to distinguish the main edges of the cube, as well as the finer edges that are inhibited on the top side of the cube. This is clearly illustrated in Fig. 9(c), where it appears to attain clearer edges when compared against the conventional edge detectors Prewitt, Sobel and Canny, which results are correspondingly shown in Figs. 9(d), 9(e) and 9(f). Supplementary videos S.5a and S.5b demonstrate the transient responses of the system as it detects the edges of the Rubik's cube shown in Fig. 9(a) and of a noise version (The input image was distorted as was done previously in Fig 7(c).) of the same, respectively.

Adaptation to light conditions. The vertebrate retina is capable of self-adapting to maintain the retinal response to visual objects approximately the same when the level of illumination changes. Here we demonstrate that our model behaves in a similar manner to its biological counterpart when subjected with distinct light conditions. Fig. 10(b) and 10(c) demonstrate that the proposed memristive network is capable of detecting the edges inhibited in the original image despite the two-times light variance in the original figure. In these examples we have manually adjusted the memristive threshold for achieving similar edge detection to the original system. When the image is brighter, the difference between a contour's pixels will be relatively higher, meaning that there will be more current flowing through the memristive devices that correspond to this edge, thus their state will be altered in a faster manner. In this case, matching the edge-detection performance of our system, as shown previously in Fig. 9(c), necessitates the use of a higher memristance threshold of $M_{TL} = 6.35k\Omega$.

On the other hand, when the image has a darker tone, the contrast between the pixels defining an edge will be less significant. In consequence, smaller potentials are established across the corresponding memristive fuses and their state will change at a slower manner. Likewise, a lower memristance threshold of $M_{TD} = 1.2k\Omega$ is required to achieve a similar performance to Fig. 9(c). Clearly, a lower threshold implies that more memristive fuses will exceed this threshold at any given time, justifying the detection of thicker edges as illustrated in Fig. 10(e). Supplementary videos S.6a and S.6b illustrate the transients of Figs. 10(b) and 10(e) respectively.

Since the relative contrast in the pixels of a "light" and "dark" tone image will be rather similar with that in the original image, if one maintains the same threshold ($M_T = 3k\Omega$) for both light conditions, the proposed system can in principle detect the same edges as previously showed in Fig. 9(c) (t=30 sec). However, in the case of a lighter environment the system will converge to a similar solution after t=21.8 sec, as shown in Fig. 10(c), while in the case of a darker environment the system will require double the time (t=44.7 sec) to converge into a similar solution (Fig. 10(f)).

**Fault Tolerance.** Biological systems depend on rather primitive elements whose properties often vary randomly. Yet, nature is capable of performing highly complex functions in a very reliable manner by employing redundancy. Similarly, solid-state devices and particularly memristive devices of deep submicron dimensions demonstrate a very poor yield. Given the fact that memristors are a disruptive technology, reliability and robustness of the devices becomes a significant burden. In this view, we extend our investigation on the effect defective devices could potentially have in our model.

We consider that for a 100% yield, all memristive fuses are reliably set with $R_{ON}$=100$\Omega$, $R_{OFF}$=16k$\Omega$ and $R_{init}$=200$\Omega$. In order to test the robustness of our system we model different yields, by assigning erratic initial states to a number of randomly selected memristive elements. This means that the conductance of these memristive elements differs from the normal one.

When a device is considered as faulty, its $R_{ON}$ could vary from 50% to 400% compared to the ideal scenario. Similarly, $R_{OFF}$ may be varied from 62.5% to 125% and $R_{init}$ could take any value from 50% up to 4000% when compared with the ideal values. Fig. 11 shows two circuit maps, where 25% (Fig. 11(a)) and 50% (Fig. 11(b)) of the total memristors in the network were randomly affected. The employed colour mapping corresponds to the randomly distributed initial states $M_{init}$ with green hexagons marking the unaffected devices and with red and blues the affected ones.

When the same optical conditions are applied as in Fig. 9(a), the OPL will produce a smoothed equivalent as shown in Fig. 12(a). When however the network's yield is set to 75% and 50%, the memristive grid acquires an uneven initial weight distribution that produces the smoothed versions shown in Figs. 12(b) and 12(c) respectively. The relative difference between the flawless and the affected memristive fuses are shown on Figs. 12(d) and 12(e) respectively. We can observe that the smoothing of the image will decrease, because of the high number of defective memristors. Yet, our edge detection method still holds and is capable of detecting most of the correct edges. Supplementary videos S.7a and S.7b demonstrate the time evolution of the network that causes the smoothing, while S.7c and S.7d illustrates dynamically the edges as detected for a yield of 75% and 50% respectively. Regardless the low yield values we tested for, the proposed hexagonal memristive network appears to be proficient in detecting the inhibited edges effectively. However, when the same conditions are employed in a rectangular memristive architecture, the results are not as encouraging as in the hexagonal topology. Besides the geometrical advantage that allows more unit cells to be tessellated per unit area, the hexagonal topology bares two extra interconnections per node, enhancing the system's redundancy. Therefore, the local averaging occurs with two more spatial partners, accounting for the introduced faults in the devices' characteristics.

**Conclusions.** We have presented a biorealistic model of the outer plexiform layer of the vertebrate retina, based on an hexagonal memristive grid implementation by exploiting the highly non-linear dynamic properties that memristors possess intrinsically. This implementation assists in minimising the overall complexity of other previously reported systems, while at the same time it achieves a local Gaussian filtering that facilitates an adaptive smoothing of both distorted and undistorted optical stimuli. Moreover, it was demonstrated that edge detection can be achieved by means of a simple memristor thresholding scheme implemented at the OPL's nodes outputs as well as through the collective evaluation of states of the memristive elements per node. Both smoothing and edge detection were assessed against distinct light conditions and it was shown that the proposed platform behaves in a similar manner as its biological counterpart. Finally, with yield being considered a very important aspect in deep sub-micron technologies and particularly practical memristive implementations, we have demonstrated that the proposed bio-inspired OPL model can effectively manage a relatively large variation in the devices' properties.

**Methods.** The decomposition of the employed images into equivalent biasing voltages, as well as the hexagonal and rectangular implementations of the memristive grids were performed in MATLAB. The constructed systems were simulated in PSPICE and relatively small networks were utilized for minimising the computation requirements. In the case of large memristive networks Biolek's memristor spice model [41] was utilized. Where for single node simulations we used Prodromakis' [42] as this model also considers the non-linear dopant kinetics of the employed elements. For ultra dense implementations the model described in [49] could also be adapted to relax the excessive computation requirements.

**Supplementary Information.** The supplementary files can be found at http://users.itk.ppke.hu/~gelan/SupplementaryFiles/.

**Acknowledgements.** The authors wish to acknowledge the *ERASMUS* scheme as well as the financial contribution of Dr. Wilf Corrigan and the CHIST-ERA net project "*Plasticity in NEUral Memrstive Architectures*".

**Author Information.** Correspondence and requests for materials should be addressed to A. Gelencsér (gelencser.andras@itk.ppke.hu) or T. Prodromakis (t.prodromakis@imperial.ac.uk).

**Figure Captions**

**Figure 1 | Conceptual representation of the mammalian retina.** To avoid confusion, only five major cell types are depicted with the main interconnection between them organised across the seven different layers. The visual information flows through the retina via bipolar cells and travels to the thalamus via the axons of ganglion cells. Lateral interconnections also exist between the bipolar and horizontal cells in the outer plexiform layer (OPL) and the bipolar and amacrine cells in the inner plexiform layer (IPL).

**Figure 2 | Outer plexifrom layer.** The OPL architecture is illustrated in **(a)** with ON and OFF bipolar cells contracted for simplicity. In the OPL the photoreceptor, horizontal and bipolar cells are interconnected through so called *triads*. The horizontal cells are interlinked through gap junctions, forming an extended lateral network. **(b)** A triad from the OPL is formed by cone pedicle, ON and OFF bipolar cell dendrites and horizontal cell dendrites or axons.

**Figure 3 | Schematic illustration of the memristive fuse.** Two identical memristors of opposing polarities are connected serially.

**Figure 4 | Close view of a single node of the proposed grid.** Voltage sources and serial resistors are representations of the output signals of photoreceptors, when subjected to an optical stimulus. Every node establishes a triad through hexagonally interconnected memristive fuses for imitating the dendrites or axons of horizontal cells, one memristor that represents the dendrite of bipolar cells, while the bias sources and resistors at the input stage denote the cone pedicle.

**Figure 5 | Effect of single-node biasing with a large contrast. (a)** depicts the static voltage distribution across the nodes of the network after t=30sec. The employed biasing scheme is depicted in the inset figure. **(b)** demonstrates the time-evolution of all memristive fuses states affiliated with this node. Distinct colours correspond to the neighbouring devices to the centre node as illustrated in the inset figure.

**Figure 6 | Simulated photoreceptor depolarisation due to distinct grayscale levels.**

**Figure 7 | Demonstration of local gaussian filtering with our proposed OPL model. (a)** is the original input image, **(b)** is the extrapolated smoothed version of (a) after t=30sec. **(c)** is a distorted version of (a) and the corresponding smoothed output after t=30sec is shown in **(d)**. **(e)** is the accentuated intensity mismatch between the smoothed outputs with and without distortion.

**Figure 8 | Edge detection I.** Detection of the edges of the utilized input image shown in **(a)**. Edge-detection was achieved via: **(b)** a bipolar threshold scheme at the output of the OPL nodes with $600\Omega \leq M_T \leq 2k\Omega$ and **(c)** a thresholding scheme at the memrstive fuses with $M_T$=1.6k$\Omega$. These results are compared against conventional edge-detection algorithms: **(d)** Prewitt, **(e)** Sobel and **(f)** Canny.

**Figure 9 | Edge detection II.** Detection of the edges of the utilised input image shown in **(a)** and the corresponding smoothed output after t=30sec is shown in **(b)**. Edge-detection was achieved by applying a thresholding scheme with $M_T$=3k$\Omega$, shown in **(c)**. This result is also compared against conventional edge-detection algorithms: **(d)** Prewitt, **(e)** Sobel and **(f)** Canny.

**Figure 10 | Evaluation of the memristive platform against distinct light conditions.** **(a)** and **(d)** are representations of the employed input image (Fig. 9(a)) in brighter and lighter tones. The detected edges for both conditions are illustrated in **(b)** and **(e)** when corresponding memristive thresholds are utilized $M_{TL}$=6.35k$\Omega$ and $M_{TD}$=1.2k$\Omega$. While **(c)** and **(f)** are the corresponding results when $M_{TL}$=3k$\Omega$ is the same for both cases, after the system has evolved for $t_L$=21.8sec and $t_D$=44.7sec.

**Figure 11 | Test conditions for evaluating the fault tolerance capacity of this model.** Shown is the distribution of memristive elements whose initial state ($M_{init}$) was altered by **(a)** 25% and **(b)** 50% of the total memristors in the network.

**Figure 12 | Smoothing and edge detection performance for varying memristor yield.** **(a)** Ideal scenario, **(b)** 75% yield and **(c)** 50% yield. **(d)** and **(e)** show the accentuated difference between the ideal and randomly affected networks. The corresponding detected edges are illustrated in **(f)**, **(g)** and **(h)** for yields of 100%, 75% and 50%.

Figure 1

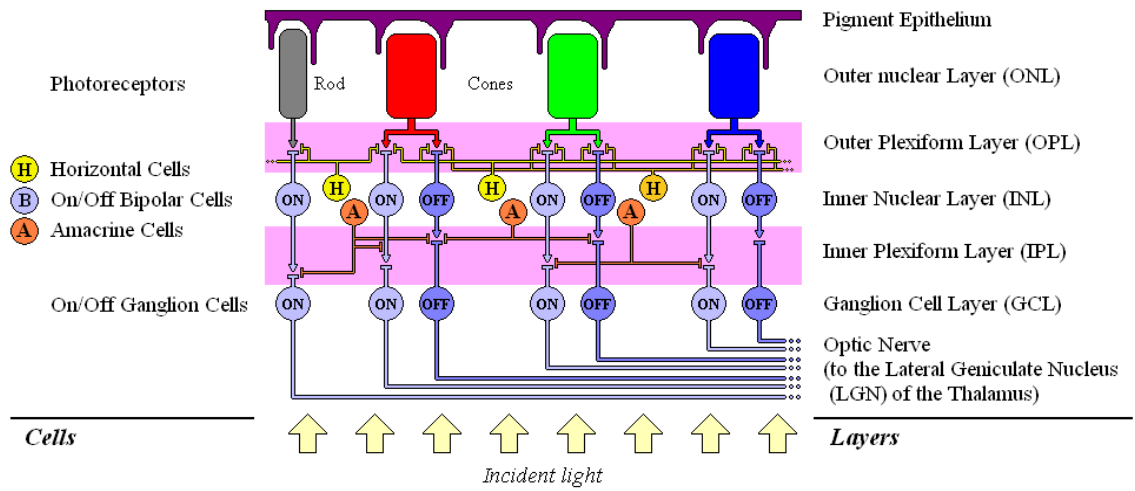

Figure 2

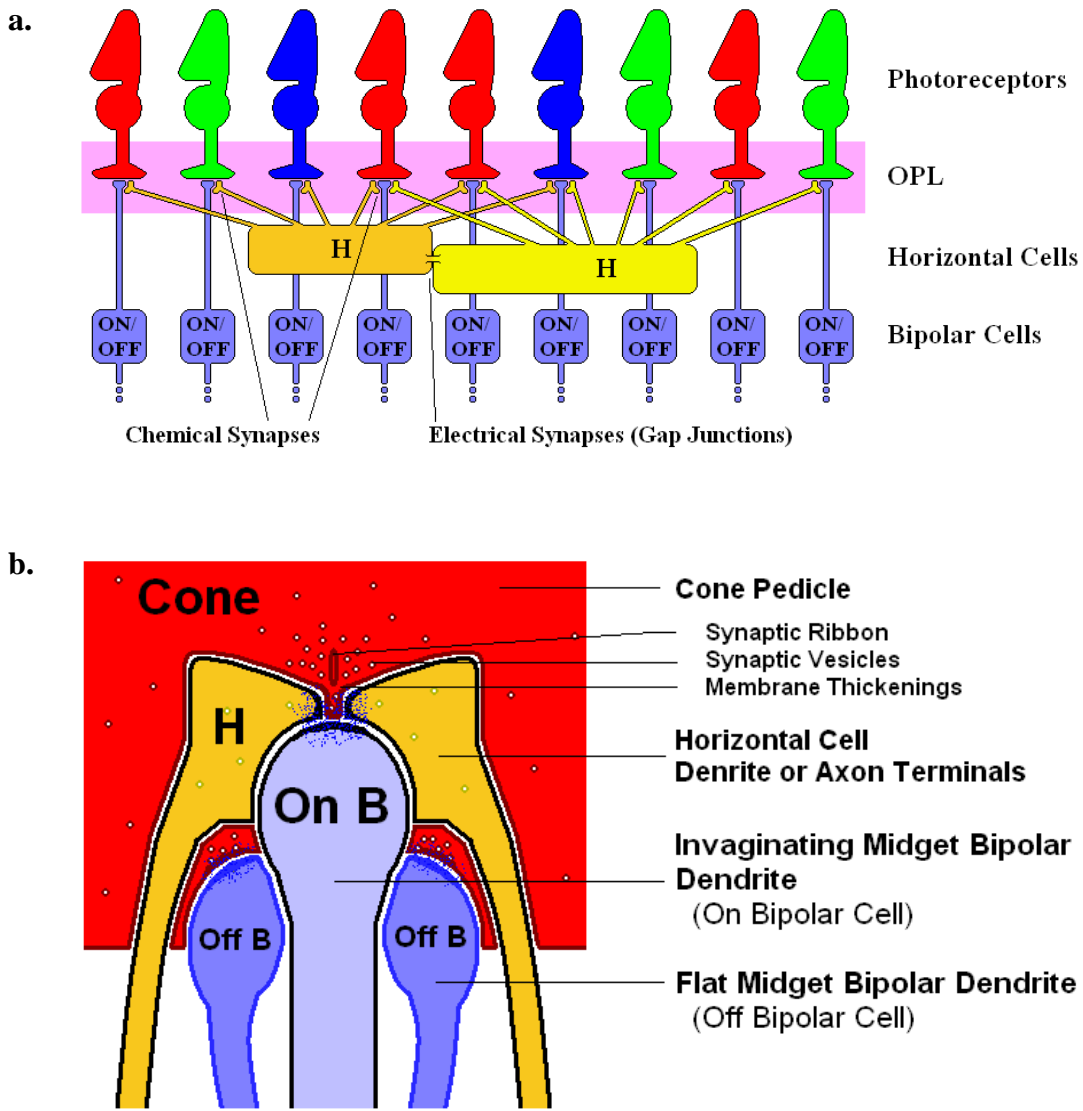

Figure 3

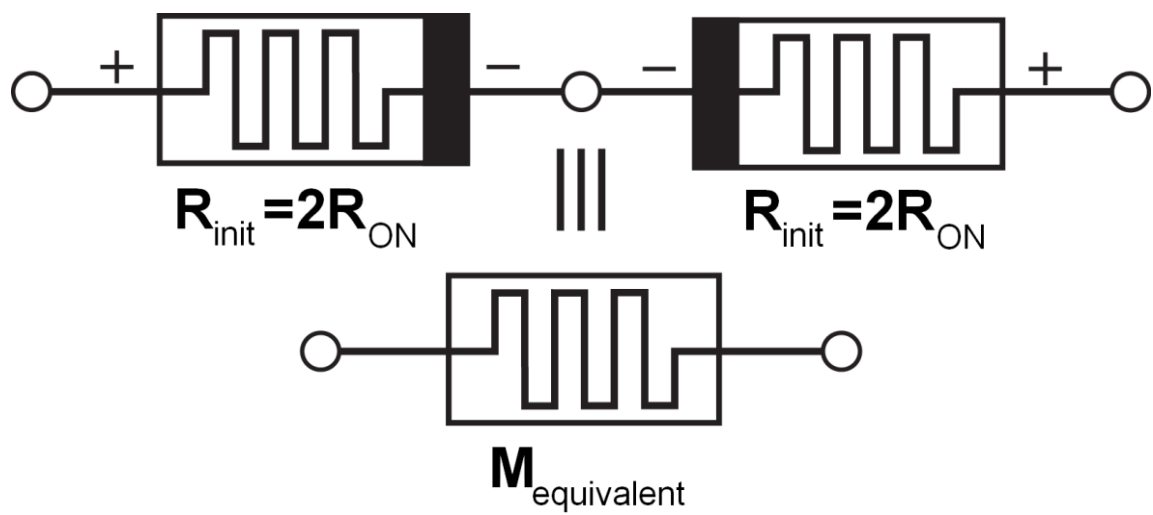

Figure 4

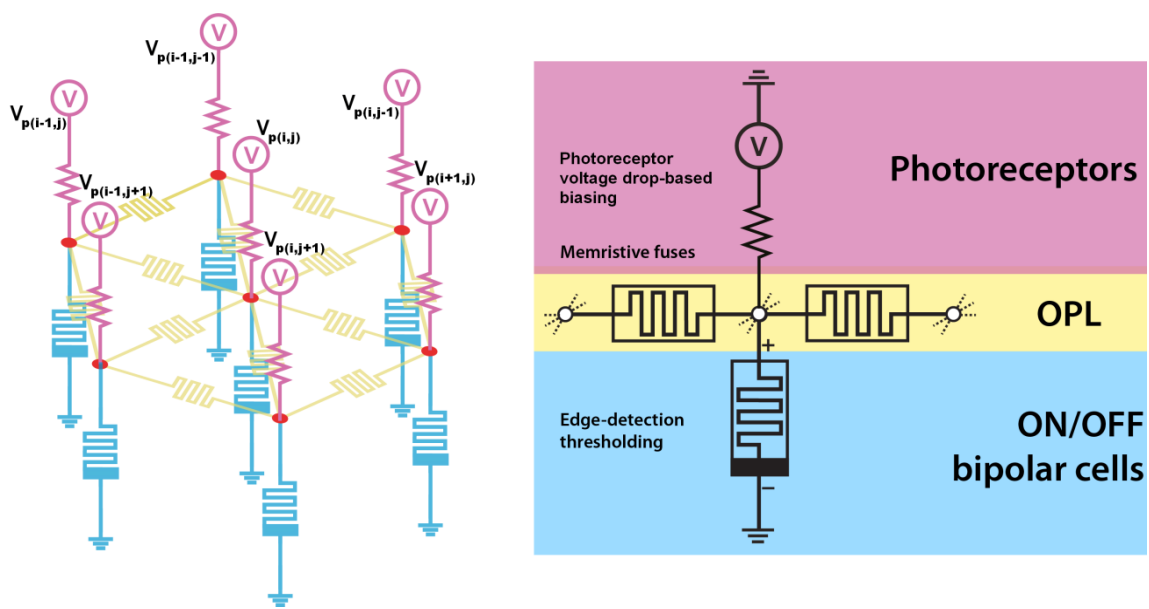

Figure 5

**a.**

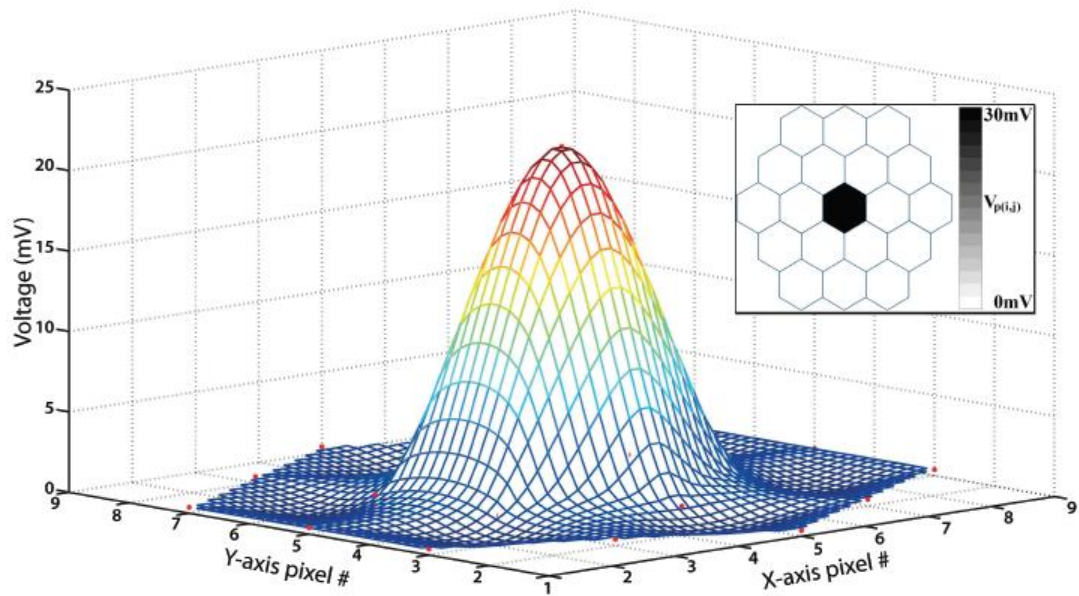

**b.**

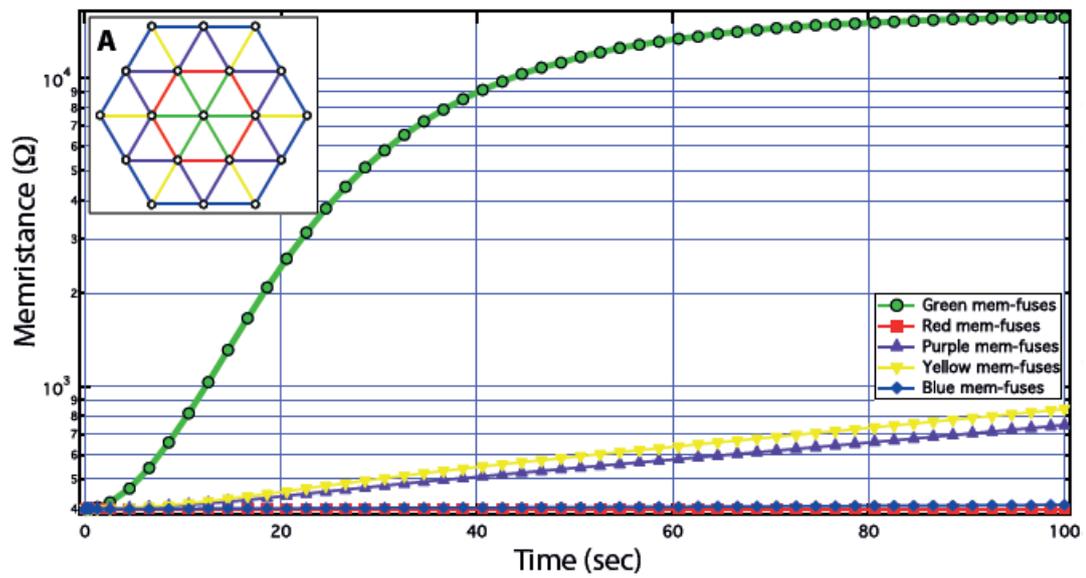

Figure 6

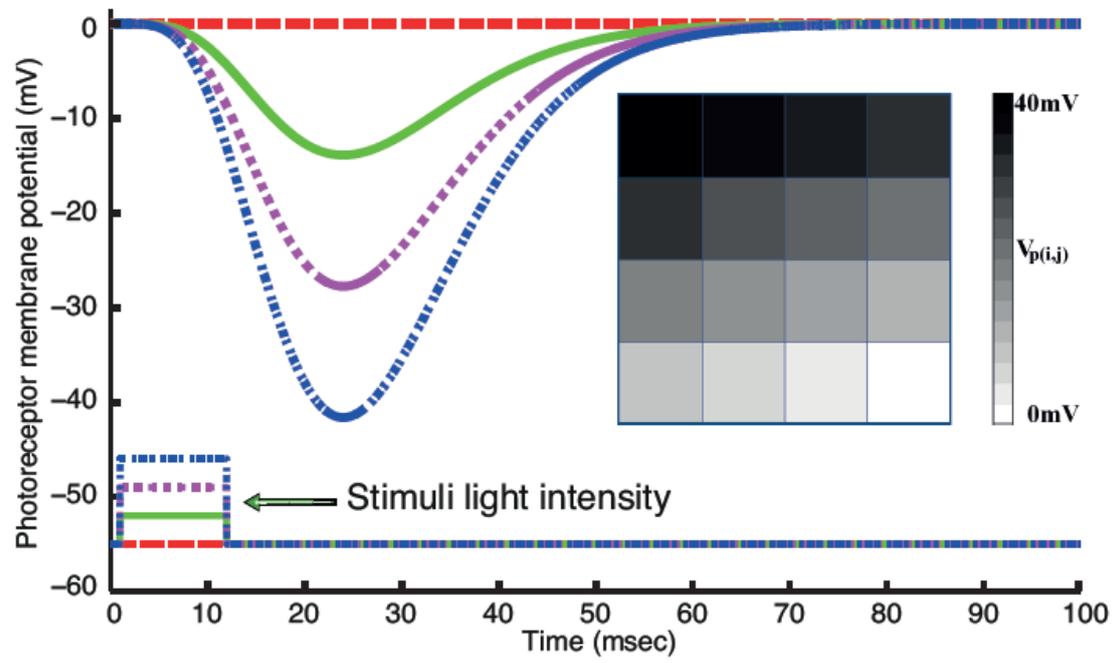

Figure 7

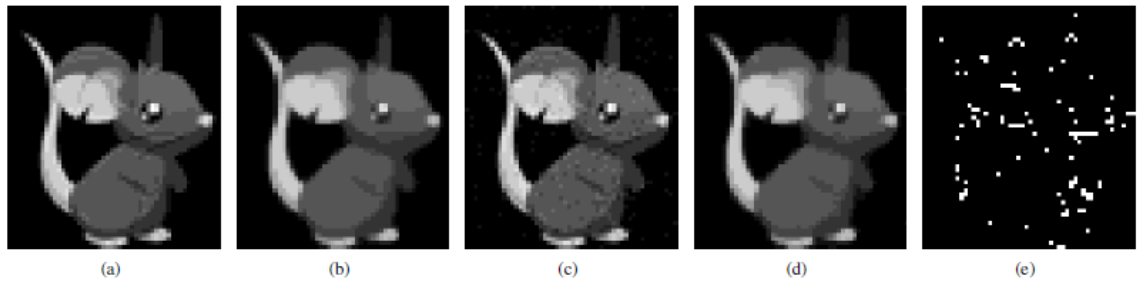

(a)     (b)     (c)     (d)     (e)

Figure 8

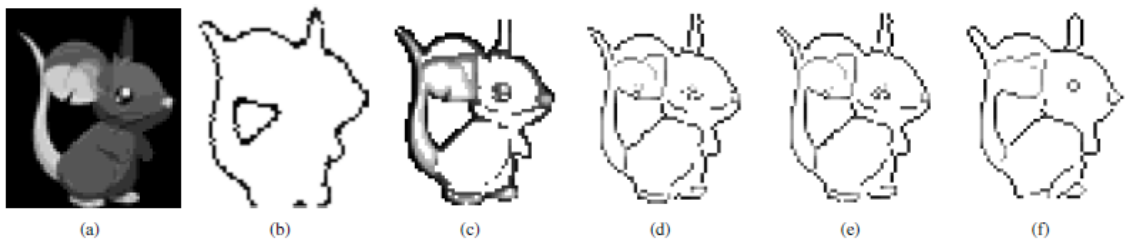

(a)     (b)     (c)     (d)     (e)     (f)

Figure 9

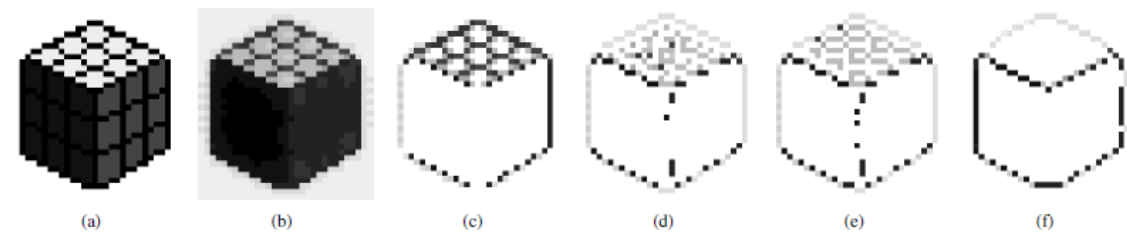

(a)     (b)     (c)     (d)     (e)     (f)

Figure 10

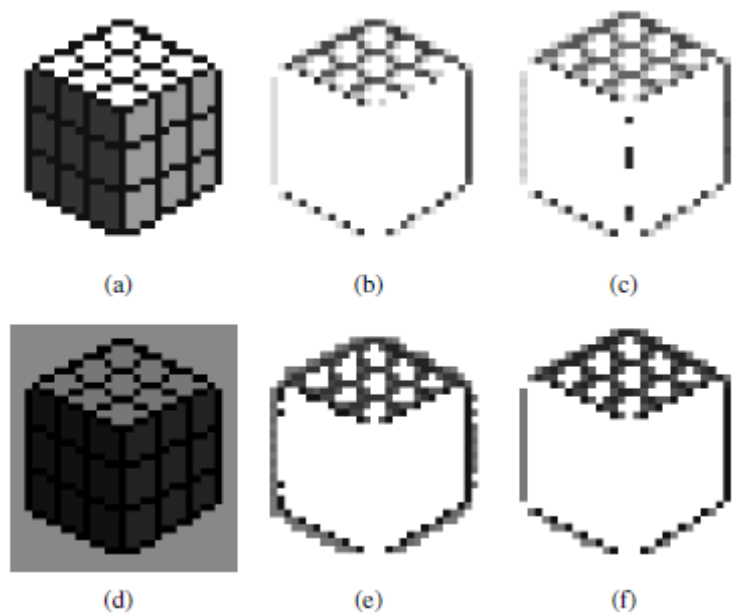

(a)     (b)     (c)

(d)     (e)     (f)

Figure 11

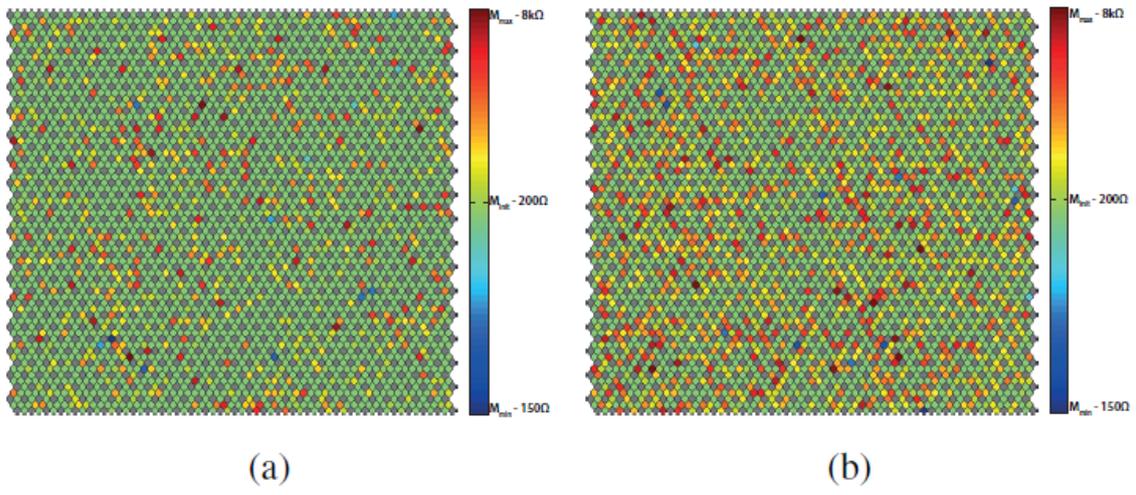

(a)  (b)

Figure 12

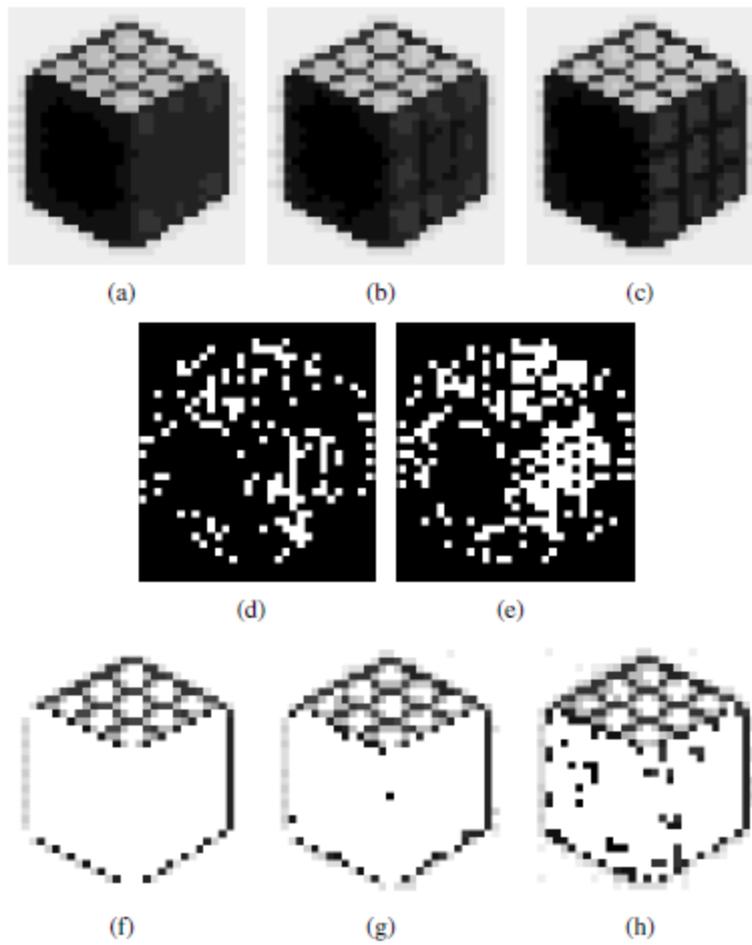

(a)  (b)  (c)

(d)  (e)

(f)  (g)  (h)